\documentclass[11pt,a4paper]{article}
\usepackage[hyperref]{naaclhlt2018}
\usepackage{times}
\usepackage{graphicx}
\usepackage{latexsym}
\usepackage{mathtools}
\usepackage{xspace}
\usepackage{algorithmicx}
\usepackage{booktabs} 
\usepackage{dblfloatfix}
\usepackage{tikz}
\usepackage{subcaption}
\usepackage{amsmath}
\usepackage{amsfonts}
\usepackage{amssymb}
\usepackage{enumitem}
\usepackage{multicol}
\usepackage{url,relsize}

\def\HS{\rule[-0.55cm]{0pt}{1.1cm}}

\DeclareCaptionFont{10pt}{\fontsize{10pt}{12pt}\selectfont}
\usepackage{url}

\aclfinalcopy 

\newcommand{\cmpclass}[1]{\textsc{#1}\xspace}
\newcommand{\ex}[1]{\textit{#1}\xspace}
\newcommand{\rile}{{RILE}\xspace}

\newcommand{\method}[1]{\textsl{#1}\xspace}
\newcommand{\Joint}{\method{Joint}}
\newcommand{\Jointstruc}{\method{Joint$_{\text{struc}}$}}
\newcommand{\Jointdoc}{\method{Joint$_{\text{doc}}$}}
\newcommand{\Jointsent}{\method{Joint$_{\text{sent}}$}}
\newcommand{\loclr}{\method{loc\_lr}}

\newcommand{\logic}[1]{\ensuremath{\mathtt{#1}}\xspace}
\newcommand{\Manifesto}{\logic{Manifesto}}
\newcommand{\Party}{\logic{Party}}
\newcommand{\RegCoalition}{\logic{RegCoalition}}
\newcommand{\EUCoalition}{\logic{EUCoalition}}
\newcommand{\SameElec}{\logic{SameElec}}
\newcommand{\Recent}{\logic{Recent}}
\newcommand{\Similarity}{\logic{Similarity}}

\newcommand{\pos}{\phantom{$-$}}

\newcommand{\secref}[2][]{Section#1~\ref{sec:#2}\xspace}
\newcommand{\tabref}[2][]{Table#1~\ref{tab:#2}\xspace}
\newcommand{\figref}[2][]{Figure#1~\ref{fig:#2}\xspace}
\newcommand{\eqnref}[2][]{Equation#1~(\ref{eq:#2})\xspace}

\title{Hierarchical Structured Model for Fine-to-coarse Manifesto Text Analysis}

\author{Shivashankar Subramanian \qquad Trevor Cohn \qquad Timothy
  Baldwin \\School of Computing and Information Systems\\ The University
  of Melbourne \\
 {\smaller\url{shivashankar@student.unimelb.edu.au} \qquad \url{{t.cohn, tbaldwin}@unimelb.edu.au}}}

\date{}

\begin{document}

\maketitle

\begin{abstract}
Election manifestos document the intentions, motives, and views of political parties.
They are often used for analysing a party's fine-grained position on a particular issue, as well as for coarse-grained positioning of a party on the left--right spectrum. 
In this paper we propose a two-stage model for automatically performing both levels of analysis over manifestos. 
In the first step we employ a hierarchical multi-task structured deep model to predict fine- and coarse-grained positions, and in the second step we perform post-hoc calibration of coarse-grained positions using probabilistic soft logic.  
We empirically show that the proposed model outperforms state-of-art approaches at both granularities using manifestos from twelve countries, written in ten different languages.
\end{abstract}

\section{Introduction}

The adoption of NLP methods has led to significant advances in the field of computational social science \cite{lazer2009life}, including political science \cite{grimmer2013text}. Among a myriad of data sources, election manifestos are a core artifact in political analysis. One of the most widely used datasets by political scientists is the Comparative Manifesto Project (CMP) dataset \cite{CMP}, which contains manifestos in various languages, covering over 1000 parties across 50 countries, from elections dating back to 1945.

In CMP, a subset of the manifestos has been manually annotated at the sentence-level with one of 57 political themes,
divided into 7 major categories.\footnote{\url{https://manifesto-project.wzb.eu/coding_schemes/mp_v5}} Such categories capture party positions (\cmpclass{favorable}, \cmpclass{unfavorable} or \cmpclass{neither}) on fine-grained policy themes, and are also useful for downstream tasks including calculating manifesto-level (policy-based) left--right position scores \cite{Budge2001,lowe2011scaling, daubler2017estimating}. An example sentence from the Green Party of England and Wales 2015 election manifesto where they take an \cmpclass{unfavorable} position on \cmpclass{military} is:
\begin{quote}
  \ex{We would: Ensure that ... less is spent on military research.}
\end{quote}
Elsewhere, they take a \cmpclass{favorable} position on \cmpclass{welfare state}:
\begin{quote}
\ex{Double Child Benefit.} 
\end{quote}
Such manual annotations are labor-intensive and prone to annotation inconsistencies \cite{coder}. In order to overcome these challenges, supervised sentence classification approaches have been proposed \cite{verberne2014automatic, ALTW2017}. 



Other than the sentence-level labels, the manifesto text also has a document-level score that quantifies its position on the left--right spectrum. Different approaches have been proposed to derive this score, based on alternate definitions of ``left--right'' \cite{slapin2008scaling, benoit2007estimating, lo2013common, daubler2017estimating}. Among these, the \rile index is the most widely adopted \cite{merz2016manifesto, oxford}, and has been shown to correlate highly with other popular scores \cite{lowe2011scaling}. 
\rile is defined as the difference between \cmpclass{right} and \cmpclass{left} positions on (pre-determined) policy themes across sentences in a manifesto \cite{budge2013}; for instance, \cmpclass{unfavorable} position on \cmpclass{military} is categorized as \cmpclass{left}. \rile is popular in CMP in particular, as mapping individual sentences to \cmpclass{left}/\cmpclass{right}/\cmpclass{neutral} categories has been shown to be less sensitive to systematic errors than other sentence-level class sets \cite{klinge, budge2013}. 

Finally, expert survey scores are gaining popularity as a means of capturing manifesto-level political positions, and are considered to be context- and time-specific, unlike \rile \cite{budge2013, daubler2017estimating}. We use the Chapel Hill Expert Survey (CHES) \cite{bakker2015measuring}, which comprises aggregated expert surveys on the ideological position of various political parties.  Although CHES is more subjective than RILE, the CHES scores are considered to be the gold-standard in the political science domain. 




In this work, we address both fine- and coarse-grained multilingual manifesto text policy position analysis, through joint modeling of sentence-level classification and document-level positioning (or ranking) tasks. We employ a two-level structured model, in which the first level captures the structure within a manifesto, and the second level captures context and temporal dependencies across manifestos.
Our contributions are as follows:
\begin{itemize}[nosep,leftmargin=1em,labelwidth=*,align=left]
\item we employ a hierarchical sequential deep model that encodes the structure in manifesto text for the sentence classification task;
\item we capture the dependency between the sentence- and document-level tasks, and also utilize additional label structure (categorization into \cmpclass{left}/\cmpclass{right}/\cmpclass{neutral}: \newcite{budge2013}) using a joint-structured model;
\item we incorporate contextual information (such as political coalitions) and encode temporal dependencies to calibrate the coarse-level manifesto position using probabilistic soft logic \cite{bach2015}, which we evaluate on the prediction of the \rile index or expert survey party position score.
\end{itemize}

\section{Related Work}
Analysing manifesto text is a relatively new application at the intersection of political science and NLP. One line of work in this space has been on sentence-level classification, including classifying each sentence according to its major political theme (1-of-7 categories) \cite{zirn2016classifying, W17-2906}, its position on various policy themes \cite{verberne2014automatic, biessmann2016automating, ALTW2017}, or its relative disagreement with other parties \cite{EMNLP2017}. Recent approaches \cite{W17-2906, ALTW2017} have also handled multilingual manifesto text (given that manifestos span multiple countries and languages; see \secref{experiment-setting}) using multilingual word embeddings.

At the document level, there has been work on using label count aggregation of (manually-annotated) fine-grained policy positions, as features for inductive analysis \cite{lowe2011scaling, daubler2017estimating}. Text-based approaches has used dictionary-based supervised methods, unsupervised factor analysis based techniques and graph propagation based approaches \cite{hjorth2015computers, 2017arXiv170704737B, EACL}. A recent paper closely aligned with our work is \newcite{ALTW2017}, who address both sentence- and document-level tasks jointly in a multilingual setting, showing that a joint approach outperforms previous approaches. But they do not exploit the structure of the text and use a much simpler model architecture: averages of word embeddings, versus our bi-LSTM encodings; and they do not leverage domain information and temporal regularities that can influence policy positions \cite{greene2016competing}. This work will act as a baseline in our experiments in \secref{experiment}.

Policy-specific position classification can be seen as related to target-specific stance classification \cite{Mohammad}, except that the target is not explicitly mentioned in most cases. Secondly, manifestos have both fine- and coarse-grained positions, similar to sentiment analysis \cite{mcdonald2007structured}. Finally, manifesto text is well structured within and across documents (based on coalition), has temporal dependencies, and is multilingual in nature.

\section{Proposed Approach}
\label{sec:approach}

In this section, we detail the first step of our two-stage approach. We use a hierarchical bidirectional long short-term memory (``bi-LSTM'') model \cite{Hochreiter:Schmidhuber:1997,bilstm, li2015hierarchical} with a multi-task objective for the sentence classification and document-level regression tasks. A post-hoc calibration of coarse-grained manifesto position is given in \secref{PSL}. 

Let $D$ be the set of manifestos, where a manifesto $d\in D$ is made up of $L$ sentences, and a sentence $s_{i}$ has $T$ words: $w_{i1}, w_{i2}, ... w_{iT}$.
The set $D_{s}\subset D$ is annotated at the sentence-level with positions on fine-grained policy issues (57 classes). The task here is to learn a model that can: (a) classify sentences according to policy issue classes; and (b) score the overall document on the policy-based left--right spectrum (\rile), in an inter-dependent fashion.

\textbf{Word encoder}: We initialize word vector representations using a multilingual word embedding matrix, $W_{e}$. We construct $W_{e}$ by aligning the embedding matrices of all the languages to English, in a pair-wise fashion. Bilingual projection matrices are built using pre-trained FastText monolingual embeddings \cite{bojanowski2016enriching} and a dictionary $D$ constructed by translating 5000 frequent English words using Google Translate. Given a pair of embedding matrices $E$ (English) and $O$ (Other), we use singular value decomposition of $O^TDE$ (which is $U \Sigma V^T$) to get the projection matrix ($W^{*}$=$UV^T$),  since it also enforces monolingual invariance \cite{artetxe2016learning, ICLR2017}. Finally, we obtain the aligned embedding matrix, $W_{e}$, as $OW^{*}$.


We use a bi-LSTM to derive a vector representation of each word in context. The bi-LSTM traverses the sentence $s_{i}$ in both the forward and backward directions, and the encoded representation for a given word $\mathbf{w}_{it} \in s_{i}$, is defined by concatenating its forward ($\overrightarrow{\mathbf{h}}_{it}$) and backward hidden states ($\overleftarrow{\mathbf{h}}_{it}$), $t \in \big[1,T\big]$. 

\textbf{Sentence model}: Similarly, we use a bi-LSTM to generate a sentence embedding from the word-level bi-LSTM, where each input sentence $s_{i}$ is represented using the last hidden state of both the forward and backward LSTMs. The sentence embedding is obtained by concatenating the hidden representations of the sentence-level bi-LSTM, in both the directions, $\mathbf{h}_{i}$ = \big[$\overrightarrow{\mathbf{h}}_{i}$, $\overleftarrow{\mathbf{h}}_{i}$\big], i $\in \big[1,L\big]$. With this representation, we perform fine-grained classification (to one-of-57 classes), using a softmax output layer for each sentence. We minimize the cross-entropy loss for this task, over the sentence-level labeled set $D_{s} \subset D$. This loss is denoted $\mathcal{L}_{S}$.

\textbf{Document model}: To represent a document $d$ we use average-pooling over the sentence representations $\mathbf{h}_{i}$ and predicted output distributions ($\mathbf{y}_i$) of individual sentences,\footnote{Preliminary experiments suggested that this representation performs better than using either hidden representations or just the output distribution.} i.e.,
$ \mathbf{V}_d = \frac{1}{L}\sum_{i \in d} \left[\begin{array}{cc} \mathbf{y}_{i} \\ \mathbf{h}_{i} \end{array} \right]  $. The range of \rile is $[-100,100]$, which we scale to the range $[-1,1]$, and model using a final $\tanh$ layer.
We minimize the mean-squared error loss function between the predicted $\hat{r}_{d}$ and actual \rile score $r_{d}$, which is denoted as $\mathcal{L}_D$:
\begin{equation}
\mathcal{L}_D = \frac{1}{|D|}  \sum_{d=1}^{|D|} \|\hat{r}_d - r_d\|^2_2 
\label{eq:doc-loss}
\end{equation}

Overall, the loss function for the joint model (\figref{HNN}), combining $\mathcal{L}_{S}$ and $\mathcal{L}_{D}$, is:
\begin{equation}
 \mathcal{L}_J = \alpha \mathcal{L}_S + (1-\alpha) \mathcal{L}_D
 \label{eq:joint-loss}
\end{equation}
where $0 \le \alpha \le 1$ is a hyper-parameter which is tuned on a development set. 

\subsection{Joint-Structured Model} 
\label{sec:SM}

The \rile score is calculated directly from the sentence labels, based on mapping each label according to its positioning on policy themes, as \cmpclass{left}, \cmpclass{right} and \cmpclass{neutral} \cite{budge2013}. Specifically, 13 out of 57 classes are categorized as \cmpclass{left}, another 13 as \cmpclass{right}, and the rest as \cmpclass{neutral}. We employ an explicit structured loss which minimizes the deviation between  sentence-level \cmpclass{left}/\cmpclass{right}/\cmpclass{neutral} polarity  predictions $\mathbf{p}$ and the document-level \rile score. The motivation to do this is two-fold: (a) enabling interaction between the  sentence- and document-level tasks with homogeneous target space (polarity and \rile); and (b) since we have more documents with just \rile and no sentence-level labels,\footnote{Strictly speaking, for these documents even, sentence annotation was used to derive the \rile score, but the sentence-level labels were never made available.} augmenting an explicit semi-supervised learning objective could propagate down the \rile label to generate sentence labels that concord with the document score.

\begin{figure}
  \begin{center}
    \includegraphics[height=7cm]{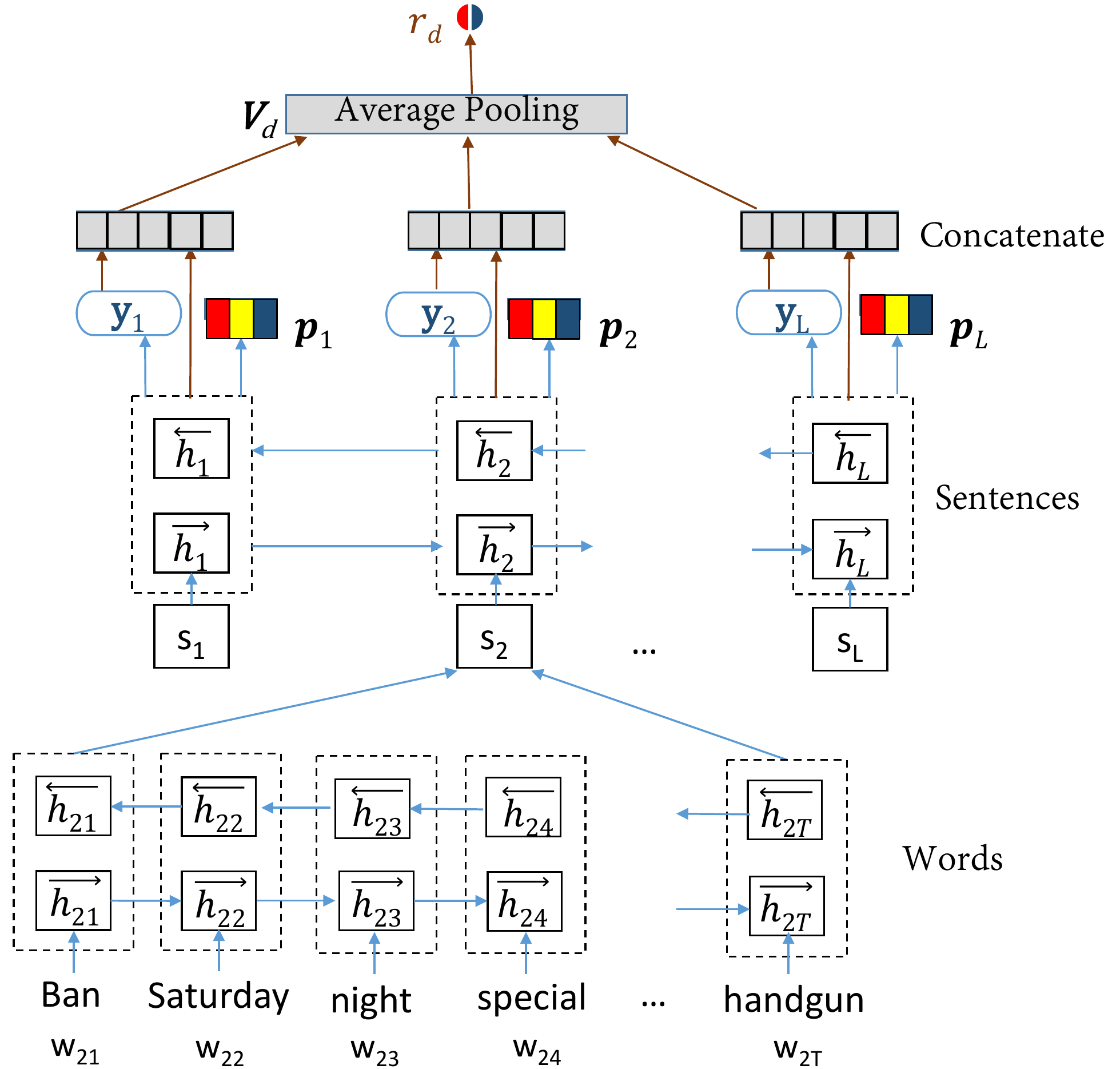}
  \end{center}
\caption{Hierarchical bi-LSTM for joint sentence--document analysis ($\mathbf{y}_{i}$ denotes the predicted 57-class distribution of sentence $s_{i}$; $\mathbf{p}_{i}$ denotes the distribution over \cmpclass{left} (in red), \cmpclass{right} (in blue) and \cmpclass{neutral} (in yellow); ${r}_{d}$ denotes the \rile score of $d$).}
\label{fig:HNN}
\end{figure}

For the sentence-level polarity prediction (shown in \figref{HNN}), we use cross-entropy loss over the sentence-level labeled set $D_{s} \subset D$, which is denoted as $\mathcal{L}_{S_P}$. 
The explicit structured sentence-document loss is given as:
  
   \begin{equation}
   \label{eq:doc-sloss}
   \begin{split}
    \mathcal{L}_{struc} = & \frac{1}{|D|} \sum_{d=1}^{|D|} \left( \frac{1}{L_d} \sum_{i \in d}(p_{{i_\mathit{right}}} - p_{{i_\mathit{left}}}) - r_{d} \right)^2
    \end{split}    
  \end{equation}    
where $p_{{i_\mathit{right}}}$ and $p_{{i_\mathit{left}}}$ are the predicted \cmpclass{right} and \cmpclass{left} class probabilities for a sentence $s_i$ ($\in d$), $r_{d}$ is the actual \rile score for the document $d$, and $L_d$ is the length of each document, d $\in$ D.
We augment the joint model's loss function (\eqnref{joint-loss}) with $\mathcal{L}_{S_P}$ and $\mathcal{L}_{struc}$ to generate a regularized multi-task loss:
\begin{equation}
 \mathcal{L}_T = \mathcal{L}_J + \beta \mathcal{L}_{S_P} + \gamma \mathcal{L}_{struc}
 \label{eq:total-loss}
\end{equation}
where $\beta, \gamma \ge 0$ are hyper-parameters which are, once again, tuned on the development set. We refer to the model trained with \eqnref{joint-loss} as ``\Joint'', and that trained with \eqnref{total-loss} as ``\Jointstruc''.

\section{Manifesto Position Re-ranking} 
\label{sec:PSL}

We leverage party-level information to enforce smoothness and regularity in manifesto positioning on the left--right spectrum \cite{greene2016competing}. For example, manifestos released by parties in a coalition are more likely to be closer in \rile score, and a party's position in an election is often a relative shift from its position in earlier election, so temporal information can provide smoother estimations. 


 \begin{table*}[!t]
 \begin{small}
  \begin{tabular}{p{15.6cm}}
 \toprule
 \textbf{PSL$_{coal}$ --- Coalition features} \\
 \midrule
 $\mathtt{Manifesto(x)} \wedge \mathtt{Party(x, a)} \wedge \mathtt{Manifesto(y)} \wedge \mathtt{Party(y, b)} \wedge \mathtt{SameElec(x, y)} \wedge \mathtt{RegCoalition(a, b)} \wedge \mathtt{pos(x)} \rightarrow \mathtt{pos(y)}$ \\
  $\mathtt{Manifesto(x)} \wedge \mathtt{Party(x, a)} \wedge \mathtt{Manifesto(y)} \wedge \mathtt{Party(y, b)} \wedge \mathtt{SameElec(x, y)} \wedge \mathtt{RegCoalition(a, b)} \wedge \neg\mathtt{pos(x)} \rightarrow \neg\mathtt{pos(y)}$ \\   
  $\mathtt{Manifesto(x)} \wedge \mathtt{Party(x, a)} \wedge \mathtt{Manifesto(y)} \wedge \mathtt{Party(y, b)} \wedge \mathtt{Recent(x, y)} \wedge \mathtt{EUCoalition(a, b)} \wedge \mathtt{pos(x)} \rightarrow \mathtt{pos(y)}$ \\
$\mathtt{Manifesto(x)} \wedge \mathtt{Party(x, a)} \wedge \mathtt{Manifesto(y)} \wedge \mathtt{Party(y, b)} \wedge \mathtt{Recent(x, y)} \wedge \mathtt{EUCoalition(a, b)} \wedge \neg\mathtt{pos(x)} \rightarrow \neg\mathtt{pos(y)}$ \\ 
 \midrule
 Transitivity \\
  \midrule
   \begin{math}
    \begin{array}{l}
      \mathtt{Manifesto(x)} \wedge \mathtt{Party(x, a)} \wedge \mathtt{Manifesto(y)} \wedge \mathtt{Party(y, b)} \wedge \mathtt{Manifesto(z)} \wedge \mathtt{Party(z, c)}\quad\wedge  \\
 \mathtt{SameElec(x, y)} \wedge \mathtt{SameElec(y, z)} \wedge \mathtt{RegCoalition(a, b)} \wedge \mathtt{RegCoalition(b, c)} \wedge \mathtt{pos(x)}
    \end{array}
  \rightarrow \text{pos(z)} 
\end{math} \\
\HS
   \begin{math}
    \begin{array}{l}
      \mathtt{Manifesto(x)} \wedge \mathtt{Party(x, a)} \wedge \mathtt{Manifesto(y)} \wedge \mathtt{Party(y, b)} \wedge \mathtt{Manifesto(z)} \wedge \mathtt{Party(z, c)}\quad\wedge  \\
 \mathtt{SameElec(x, y)} \wedge \mathtt{SameElec(y, z)} \wedge \mathtt{RegCoalition(a, b)} \wedge \mathtt{RegCoalition(b, c)} \wedge \neg\mathtt{pos(x)}
    \end{array}
  \rightarrow \neg\mathtt{pos(z)} 
\end{math} \\
\begin{math}
  \begin{array}{l}
      \mathtt{Manifesto(x)} \wedge \mathtt{Party(x, a)} \wedge \mathtt{Manifesto(y)} \wedge \mathtt{Party(y, b)} \wedge \mathtt{Manifesto(z)} \wedge \mathtt{Party(z, c)} \quad\wedge  \\
 \mathtt{Recent(x, y)} \wedge \mathtt{Recent(y, z)} \wedge 
\mathtt{EUCoalition(a, b)} \wedge \mathtt{EUCoalition(b, c)} \wedge \mathtt{pos(x)}
    \end{array}
  \rightarrow \text{pos(z)} 
\end{math} \\
\HS
   \begin{math}
    \begin{array}{l}
      \mathtt{Manifesto(x)} \wedge \mathtt{Party(x, a)} \wedge \mathtt{Manifesto(y)} \wedge \mathtt{Party(y, b)} \wedge \mathtt{Manifesto(z)} \wedge \mathtt{Party(z, c)} \quad\wedge  \\
 \mathtt{Recent(x, y)} \wedge \mathtt{Recent(y, z)} \wedge 
\mathtt{EUCoalition(a, b)} \wedge \mathtt{EUCoalition(b, c)} \wedge \neg\mathtt{pos(x)}
    \end{array}
  \rightarrow \neg\mathtt{pos(z)} 
\end{math} \\
\midrule
\midrule        
 \textbf{PSL$_{esim}$ --- Similarity-based relational feature} \\
 \midrule
    $\mathtt{Manifesto(x)} \wedge \mathtt{Manifesto(y)} \wedge \mathtt{Similarity(x, y)} \wedge \mathtt{Recent(x, y)} \wedge \mathtt{pos(x)} \rightarrow \mathtt{pos(y)}$ \\
  $\mathtt{Manifesto(x)} \wedge \mathtt{Manifesto(y)} \wedge \mathtt{Similarity(x, y)} \wedge \mathtt{Recent(x, y)} \wedge \neg\mathtt{pos(x)} \rightarrow \neg\mathtt{pos(y)}$ \\  \midrule
  \midrule        
  \textbf{PSL$_{ploc}$ --- Right--left ratio} \\
  \midrule
    $\mathtt{Manifesto(x)} \wedge \mathtt{LwRightLeftRatio(x)} \rightarrow \mathtt{pos(x)}$ \\
        $\mathtt{Manifesto(x)} \wedge \neg \mathtt{LwRightLeftRatio(x)} \rightarrow \neg\mathtt{pos(x)}$ \\
   \midrule 
   \midrule        
  \textbf{PSL$_{temp}$--- Temporal Dependency} \\
   \midrule
   $\mathtt{Manifesto(x)} \wedge \mathtt{Party(x, a)} \wedge \mathtt{PreviousManifesto(x, a, t)} \wedge \mathtt{pos(t)} \rightarrow \mathtt{pos(x)}$ \\    
 $\mathtt{Manifesto(x)} \wedge \mathtt{Party(x, a)} \wedge \mathtt{PreviousManifesto(x, a, t)} \wedge \neg\mathtt{pos(t)} \rightarrow \neg\mathtt{pos(x)}$ \\    
 \bottomrule
  \end{tabular}
  \caption{PSL Model: Values for \Similarity, \logic{LwRightLeftRatio} and \logic{pos} are obtained from the joint-structured model (Figure \ref{fig:HNN}). Except for \logic{pos}, other values are fixed in the network. Domain (\logic{y}) for \logic{SameElec (x, y)} is within the country, and for \logic{Recent(x, y)} covers all the countries. $\neg$ denotes negation. Distance to satisfaction for each ground rule is obtained using a hinge-loss potential, which is then used inside the HL-MRF model (\eqnref{HL-MRF}), where \logic{pos} is $\mathbf{Y}$.}
  \label{tab:pslrule}
  \end{small}
\end{table*}

\subsection{Probabilistic Soft Logic}

To address this, we propose an approach using hinge-loss Markov random fields (``HL-MRFs''), a scalable class of continuous, conditional graphical models \cite{bach2013}. HL-MRFs have been used for many tasks including political framing analysis on Twitter \cite{johnson2017leveraging} and user stance classification on socio-political issues \cite{sridhar2014collective}. These models can be specified using Probabilistic Soft Logic (``PSL'') \cite{bach2015}, a weighted first order logical template language. An example of a PSL rule is
\[ \lambda: \mathtt{P(a)} \wedge \mathtt{Q(a, b)} \rightarrow \mathtt{R(b)} \]
where $\mathtt{P}$, $\mathtt{Q}$, and $\mathtt{R}$ are predicates, $\mathtt{a}$ and $\mathtt{b}$ are variables, and $\lambda$ is the weight associated with the rule. PSL uses soft truth values for predicates in the interval $\big[0,1\big]$. The degree of ground rule satisfaction is determined using the Lukasiewicz t-norm and its corresponding co-norm as the relaxation of the logical AND and OR, respectively. The weight of the rule indicates its importance in the HL-MRF probabilistic model, which defines a probability density function of the form:
\begin{equation}
\begin{split}
  P(\mathbf{Y}|\mathbf{X}) & \propto \exp \left(-\sum_{r=1}^{M} \lambda_{r} \phi_{r}(\mathbf{Y}, \mathbf{X}) \right), \\
  \phi_r(\mathbf{Y}, \mathbf{X}) & = \max{\{l_{r}(\mathbf{Y}, \mathbf{X}), 0 \}}^{\rho_{r}},
\end{split}
\label{eq:HL-MRF}
\end{equation}
where $\phi_{r}$($\mathbf{Y}, \mathbf{X}$) is a hinge-loss potential corresponding to an instantiation of a rule, and is specified by a linear function $l_{r}$ and optional exponent $\rho_r$ $\in \{1, 2\}$. Note that the hinge-loss potential captures the distance to satisfaction.\footnote{Degree of satisfaction for the example PSL rule $r$, $\neg \mathtt{P} \vee \neg \mathtt{Q} \vee \mathtt{R}$, using the Lukasiewicz co-norm is given as $\min\{2 - \mathtt{P} - \mathtt{Q} + \mathtt{R}, 1\}$. From this, the distance to satisfaction is given as $\max\{\mathtt{P} + \mathtt{Q} - \mathtt{R} - 1, 0\}$, where $\mathtt{P} + \mathtt{Q} -\mathtt{R} - 1$ indicates the linear function $l_{r}$.}

\subsection{PSL Model} 
\label{PSLModel}

Here we elaborate our PSL model (given in \tabref{pslrule}) based on coalition information, manifesto content-based features (manifesto similarity and right--left ratio), and temporal dependency. Our target \logic{pos} (calibrated \rile) is a continuous variable $\big[0,1\big]$, where 1 indicates that a manifesto occupies an extreme right position, 0 denotes an extreme left position, and 0.5 indicates center. Each instance of a manifesto and its party affiliation are denoted by the predicates \Manifesto and \Party.

\begin{description}
\item \textbf{Coalition}: We model multi-relational networks based on regional coalitions within a given country (\RegCoalition),\footnote{\url{http://www.parlgov.org/}} and also cross-country coalitions in the European parliament (\EUCoalition).\footnote{\url{http://www.europarl.europa.eu}} We set the scope of interaction between manifestos (\logic{x} and \logic{y}) from a country to the same election (\SameElec). For manifestos across countries, we consider only the most recent manifesto (\logic{Recent}) from each party (\logic{y}), released within 4 years relative to \logic{x}. We use a logistic transformation of the number of times two parties have been in a coalition in the past (to get a value between 0 and 1), for both \RegCoalition and \EUCoalition. We also construct rules based on transitivity for both the relational features, i.e., parties which have had common coalition partners, even if they were not allies themselves, are likely to have similar policy positions.

\item \textbf{Manifesto similarity}: Manifestos that are similar in content are expected to have similar \rile scores (and associated sentence-level label distributions), similar to the modeling intuition captured by \newcite{Burford+:2015} in the context of congressional debate vote prediction.  For a pair of recent manifestos (\Recent) we use the cosine similarity (\Similarity) between their respective document vectors $\mathbf{V_d}$ (Figure \ref{fig:HNN}).

\item \textbf{Right--left ratio}: For a given manifesto, we compute the ratio of sentences categorized under \cmpclass{right} to \cmpclass{others} ($\frac{\cmpclass{\# right}}{\cmpclass{\# right}+\cmpclass{\# left}+\cmpclass{\# neutral}})$, where the categorization for sentences is obtained using the joint-structured model (\eqnref{total-loss}). We also encode the location of sentence $l_{s}$ in a document, by weighing the count of sentences for each class $C$ by its location value $\sum_{s\in C}\log(l_{s}+1)$ (referred to as \loclr). The intuition here is that the beginning parts of a manifesto tends to contain generic information such as preamble, compared to later parts which are more policy-dense.  We perform a logistic transformation of \loclr to derive the \logic{LwRightLeftRatio}. 

\item \textbf{Temporal dependency}: We capture the temporal dependency between a party's current manifesto position and its previous manifesto position (\logic{PreviousManifesto}).
\end{description}

Other than for the look-up based random variables, the network is instantiated with predictions (for \Similarity, \logic{LwRightLeftRatio} and \logic{pos}) from the joint-structured model (Figure \ref{fig:HNN}). All the random variables, except \logic{pos} (which is the target variable), are fixed in the network. These values are then used inside a PSL model for collective probabilistic reasoning, where the first-order logic given in Table \ref{tab:pslrule} is used to define the graphical model (HL-MRF) over the random variables detailed above. Inference on the HL-MRF is used to obtain the most probable interpretation such that it satisfies most ground rule instances, i.e., considering the relational and temporal dependencies.

\section{Evaluation}
\label{sec:experiment}

\subsection{Experimental Setup}
\label{sec:experiment-setting}

As our dataset, we use manifestos from CMP for European countries only, as in \secref{psl-results} we will validate the manifesto's overall position on the left-right spectrum, using the Chapel Hill Expert Survey (CHES), which is only available for European countries \cite{bakker2015measuring}. In this, we sample 1004 manifestos from 12 European countries, written in 10 different languages --- Danish (Denmark), Dutch (Netherlands), English (Ireland, United Kingdom), Finnish (Finland), French (France), German (Austria, Germany), Italian (Italy), Portuguese (Portugal), Spanish (Spain), and Swedish (Sweden). Out of the 1004 manifestos, 272 are annotated with both sentence-level labels and \rile scores, and the remainder only have \rile scores (see \tabref{al} for further statistics). 

There are (less) scenarios where a natural sentence is segmented into sub-sentences and annotated with different classes \cite{daubler2012natural}. Hence we use NLTK sentence tokenizer followed by heuristics from \newcite{daubler2012natural} to obtain sub-sentences. Consistent with previous work \cite{ALTW2017}, we present results with manually segmented and annotated test documents.

\begin{table} [t]
\centering 
  \small
  \begin{tabular}{@{}crr} 
  \toprule
    Lang. & \# Docs (Anntd.) & \# Sents (Anntd.)\\
    \midrule
    Danish  & 175 \,\,\,(36)  &  29694 \,\,\,\,\,(8762)	\\
    Dutch  & 107 \,\,\,(48)  &  132524 \,\,(70559)	\\
    English   &  117 \,\,\,(27)& 86603 \,\,(34512) 	 \\    	
    Finnish  &  97 \,\,\,(16) &  17979 \,\,\,\,\,(8503) \\
    French    & 53 \,\,\,(10) & 22747 \,\,\,\,\,(5559)\\
    German    & 117 \,\,\,(46) & 111376 \,\,(73652) \\
    Italian    & 98 \,\,\,(15)  & 41455 \,\,\,\,\,(5154)\\
    Portuguese    & 60 \,\,\,\,\,\,(9)  & 40922 \,\,(11077)\\
    Spanish    & 85 \,\,\,(50)  & 145355 \,\,(93964)\\
    Swedish    & 95 \,\,\,(15)  & 19551 \,\,\,\,\,(7938)\\
\midrule
 Total    & 1004 (272)  & 648206 (319680)\\
    \bottomrule

  \end{tabular}
  \caption{Statistics of dataset (``Anntd.'' refers to the number of documents with sentence annotations in the second column, and the number of sentences with annotations in the third column).}
  \label{tab:al}
\end{table}

\subsection{Baseline Approaches}

Sentence-level baseline approaches include:
\begin{itemize}
\setlength\itemsep{0.15em}
\item  \textbf{\method{BoW-NN}}: TF-IDF-weighted unigram bag-of-words representation of sentences \cite{biessmann2016automating}, and monolingual training using a multi-layer perceptron (``MLP'') model.
\item \textbf{\method{BoT-NN}}: Similar to above, but trigram bag-of-words.
\item \textbf{\method{AE-NN}}: MLP model with average multilingual word embeddings as the sentence representation \cite{ALTW2017}. 
\item \textbf{\method{CNN}}: Convolutional neural network (``CNN'': \newcite{W17-2906}) with multilingual word embeddings.
\item \textbf{\method{Bi-LSTM}}: Simple bi-LSTM over multilingual word embeddings, last hidden units are concatenated to form the sentence representation, and fed directly into a softmax sentence-level layer. We evaluate two scenarios: (1) with a trainable embedding matrix $W_{e}$ (\textbf{\method{Bi-LSTM(+up)}}); and (2) without a trainable $W_{e}$.
\end{itemize}

Document-level baseline approaches include: 
\begin{itemize}
\setlength\itemsep{0.15em}
\item \textbf{\method{BoC}}: Bag-of-centroids (BoC) document representation based on clustering the word embeddings \cite{lebret2014n}, fed into a neural network regression model. 
\item \textbf{\method{HCNN}}: Hierarchical CNN, where we encode both the sentence and document using stacked CNN layers. 
\item \textbf{\method{HNN}}: State-of-the-art hierarchical neural network model of \newcite{ALTW2017}, based on average embedding representations for sentences and the document. 
\end{itemize}

We present results evaluated under two different settings: (a) 80--20\% random split averaged across 10 runs to validate the hierarchical model (\secref{random1} and \secref{random2}); and (b) temporal setting, where train- and test-set are split chronologically, to validate both the hierarchical deep model and the PSL approach especially, since we encode temporal dependencies (\secref{psl-results}).

\subsection{Hierarchical Sentence- and Document-level Model}
\label{sec:random1}

\begin{table*}[!ht]
  \centering
  \begin{small}
  \begin{tabular}{ c@{\hskip 0.2in} c c c c c c c c c}
  \toprule
    Lang. & \method{BoW-NN} & \method{BoT-NN} & \method{AE-NN} & \method{CNN} & \method{Bi-LSTM} & \method{Bi-LSTM(+up)} &\Jointsent & \Joint &  \Jointstruc \\
    \midrule
    Danish  & 	0.35 & 0.33 & 0.35 & 0.31 & 0.38 & 0.38 & \textbf{0.44} & 0.40 & 0.43\\
    Dutch   &  0.41 & 0.41 & 0.40 & 0.34 & 0.39 & 0.43 & \textbf{0.52} & 0.50 & 0.50\\    
   English   &  0.39 & 0.43 & 0.43 & 0.40 & 0.45 & 0.47 & 0.49 & \textbf{0.50} & 0.49\\     
   Finnish  &   0.30 & 0.34 & 0.33 & 0.30 & 0.38 & 0.39 & \textbf{0.44} & 0.41 & 0.42\\
    French    & 0.36 & 0.37 & 0.36 & 0.37 & 0.42 & 0.44 & 0.48 & \textbf{0.49} & 0.48 \\
    German    &  0.33 & 0.35 & 0.37 & 0.35 & 0.40 & 0.41 & 0.45 & 0.45 & \textbf{0.46}\\
    Italian    & 0.33 & 0.38 & 0.37 & 0.31 & 0.37 & 0.39 & 0.49 & \textbf{0.52} & \textbf{0.52}\\
    Portuguese    &  0.32 & 0.38 & 0.31 & 0.28 & 0.43 & \textbf{0.46} & 0.44 & 0.44& 0.43\\
    Spanish    & 0.38 & 0.39 & 0.39 & 0.35 & 0.42 & 0.41 & \textbf{0.50} & 0.49 & \textbf{0.50}\\
	Swedish    &  0.46 & 0.42 & 0.36 & 0.36 & 0.41 & 0.44 & \textbf{0.49} & 0.46  & 0.46\\    
\midrule
Avg.    & 0.36 & 0.38 & 0.38 & 0.35 & 0.40 & 0.42 & \textbf{0.48} & 0.47 & \textbf{0.48}\\

 \bottomrule
   \end{tabular}
  \end{small}
  \caption{Micro-Averaged F-measure for sentence classification. Best scores are given in bold.}
  \label{tab:sr}
\end{table*}

\begin{figure}[t]
\begin{subfigure}{0.56\textwidth}
\centering
\includegraphics[width=0.61\linewidth]{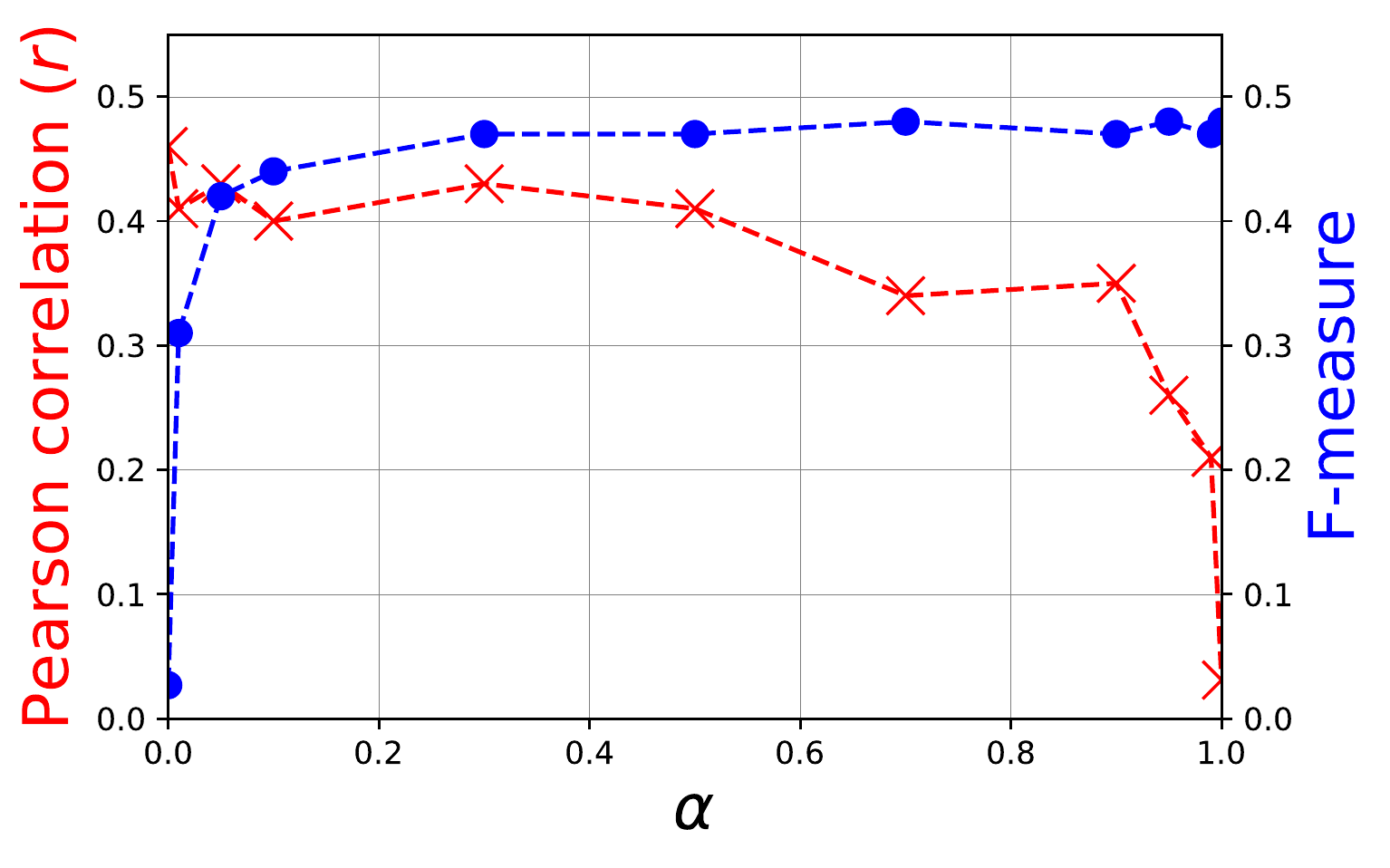}
\caption{Effect of $\alpha$ in equation \ref{eq:joint-loss}.}
\label{fig:alpha}
\end{subfigure} \hspace{0 \textwidth}

\begin{subfigure}{0.56\textwidth}
\centering
\includegraphics[width=0.61\linewidth]{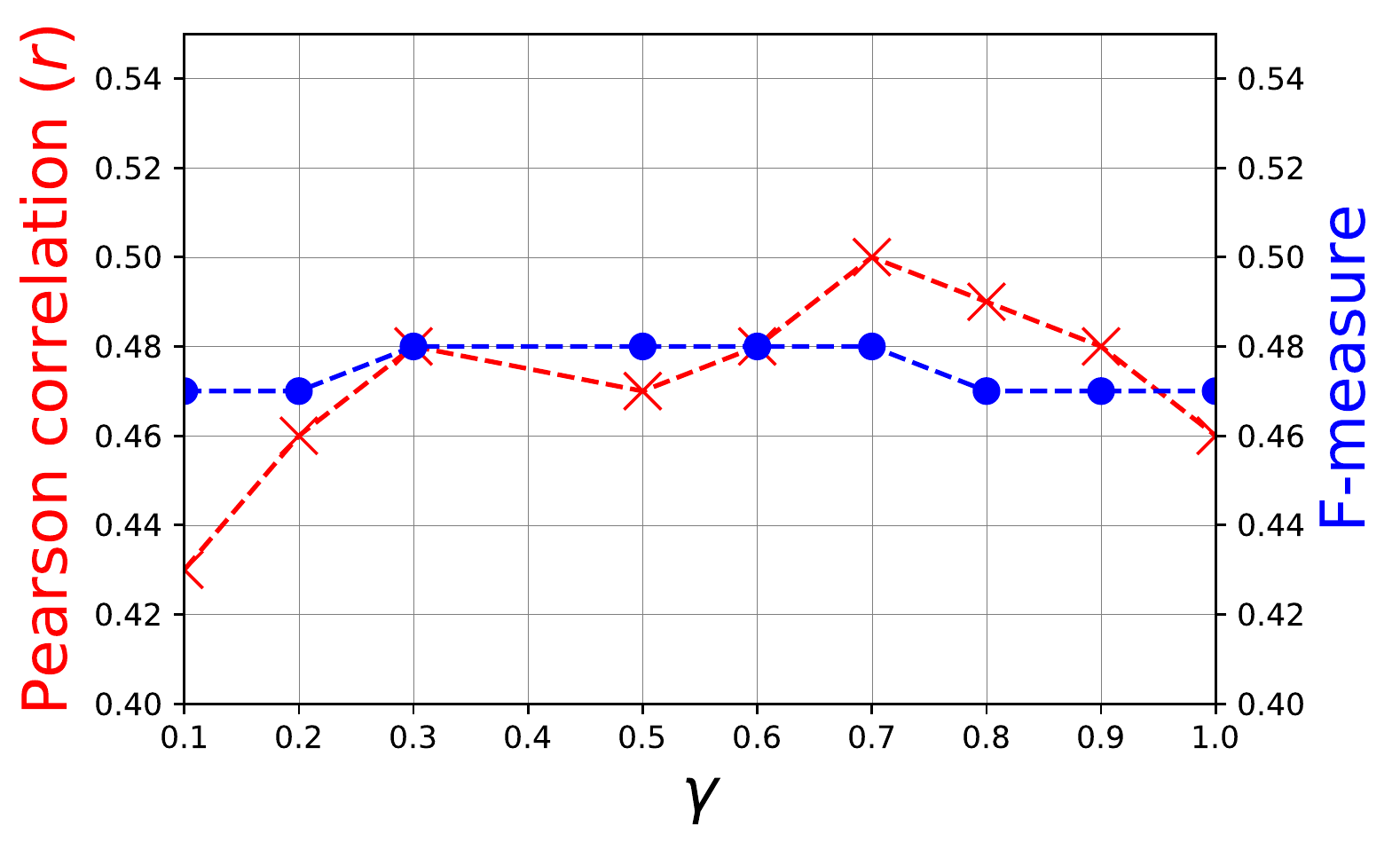}
\caption{Effect of $\gamma$ in equation \ref{eq:total-loss}.}
\label{fig:gamma}
\end{subfigure}

\begin{subfigure}{0.56\textwidth}
\centering
\includegraphics[width=0.61\linewidth]{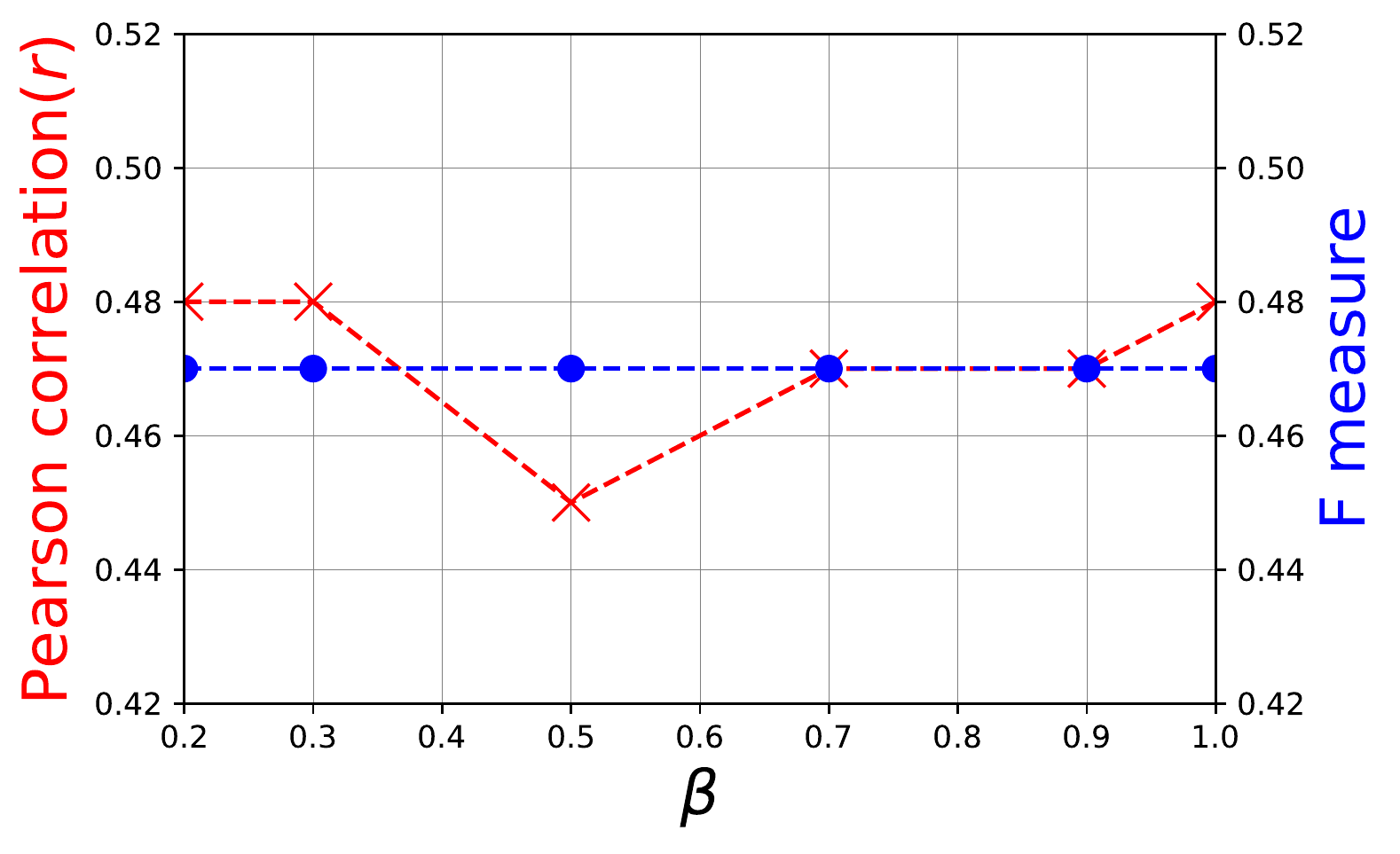}
\caption{Effect of $\beta$ in equation \ref{eq:total-loss}.}
\label{fig:beta}
\end{subfigure}

\label{annotation}
\caption{Effect of hyper-parameters on sentence- and document-level performance. \tikz\draw[blue,fill=blue] (0,0) circle (.4ex); denotes F-measure (right axis) and \color{red}$\times$ \color{black}denotes Pearson correlation (left axis).}
\end{figure}

We present sentence-level results with a 80--20\% random split in \tabref{sr}, stratified by country, averaged across 10 runs. For \method{Bi-LSTM}, we found the setting with a trainable embedding matrix (\method{Bi-LSTM(+up)}) to perform better than the non-trainable case (\method{Bi-LSTM}). Hence we use a similar setting for \Joint and \Jointstruc. We show the effect of $\alpha$ (from \eqnref{joint-loss}) in \figref{alpha}, based on which we set $\alpha=0.3$ hereafter. With the chosen model, we study the effect of the structured loss (\eqnref{total-loss}), by varying $\gamma$ with fixed  $\beta= 0.1$, as shown in \figref{gamma}. We observe that $\gamma=0.7$ gives the best performance, and varying $\beta$ with $\gamma$ at 0.7 does not result in any further improvement (see \figref{beta}). Sentence-level results measured using F-measure, for baseline approaches and the proposed models selected from \figref{alpha} (\Joint), \figref[s]{gamma} and \ref{fig:beta} (\Jointstruc) are given in \tabref{sr}. We also evaluate the special case of $\alpha=1$, in the form of sentence-only model \Jointsent. For the document-level task, results for overall manifesto positioning measured using Pearson's correlation ($r$) and Spearman's rank correlation ($\rho$) are given in \tabref{dr}. We also evaluate the hierarchical bi-LSTM model with document-level objective only, \Jointdoc.

 \begin{table}[!t]
  \centering
  \begin{small}
  \begin{tabular}{ c c c }
  \toprule
    Approach & $r$ & $\rho$\\
    \midrule
    \method{BoC}    &  0.18 & 0.20\\
    \method{HCNN}  &  0.24 & 0.26 \\
    \method{HNN} &  0.28 & 0.32 \\
    \Jointdoc &  0.30 &  0.37 \\    
    \Joint &  0.46 & 0.54 \\    
    \Jointstruc &  \textbf{0.50} & \textbf{0.63} \\    
    \bottomrule
  \end{tabular}
  \end{small}
  \caption{\rile score prediction performance. Best scores are given in bold.}
 \label{tab:dr}
\end{table}
We observe that hierarchical modeling (\Jointsent, \Joint and \Jointstruc) gives the best performance for  sentence-level classification for all the languages except Portuguese, on which it performs slightly worse than \method{Bi-LSTM(+up)}. Also, \Jointstruc, does not improve over \Jointsent. We perform further analysis to see the effect of joint-structured model on the sentence-level task under sparsely-labeled conditions in \secref{random2}. On the other hand, for the document-level task, the joint model (\Joint) performs better than  \Jointdoc and all the baseline approaches. Lastly, the joint-structured model (\Jointstruc) provides further improvement over \Joint.

 
\subsection{Analysis of Joint-Structured Model for Sentence-level task}
\label{sec:random2}
\begin{figure}
\centering
\includegraphics[scale=0.4]{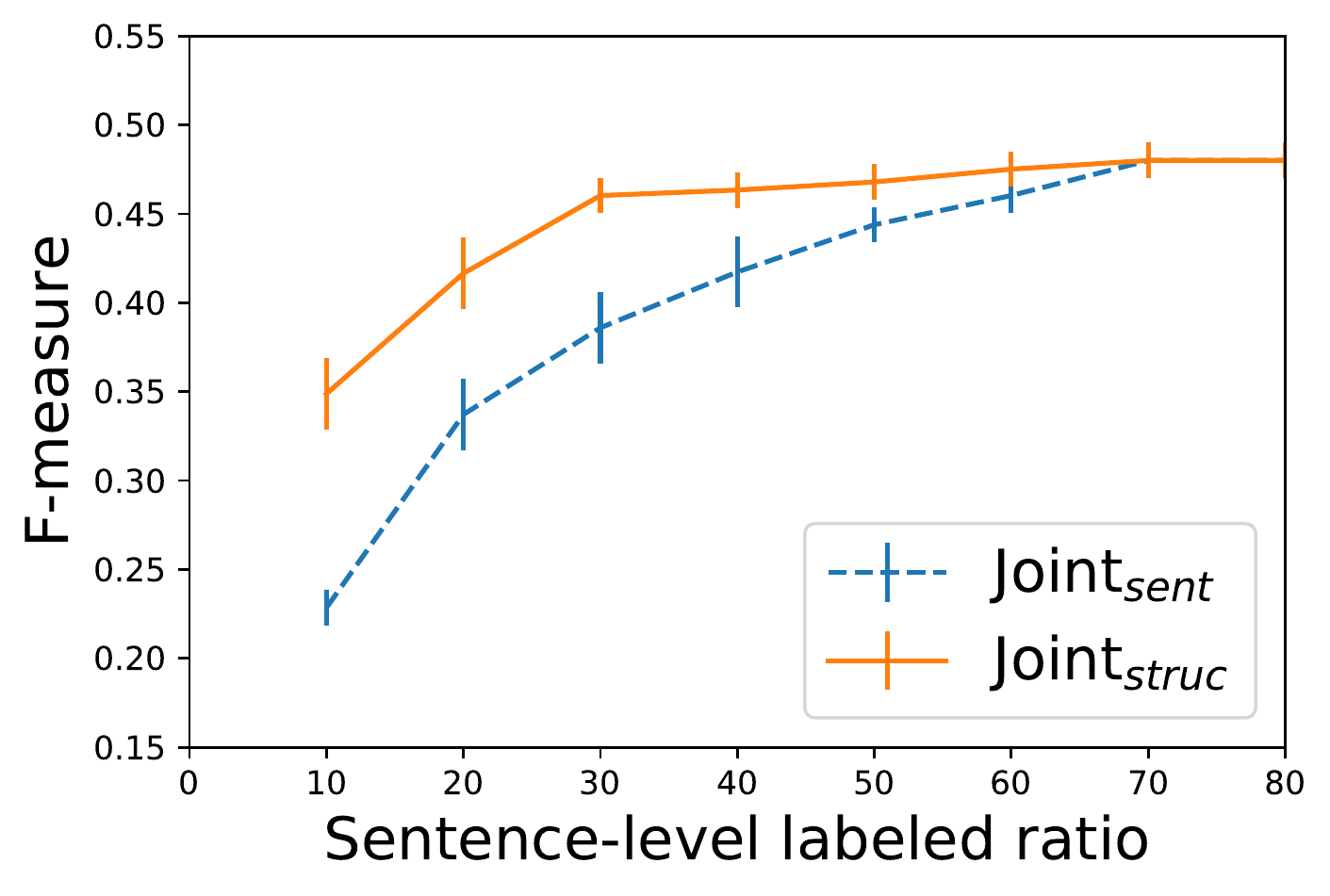}
\caption{F-measure for \Jointstruc vs.\ \Jointsent across different ratios of sentence-level labeled manifestos (averaged over 10 runs, with standard deviation)}
\label{fig:JSoverJoint}
\end{figure}

To understand the utility of joint modeling, especially given that there are more manifestos with document-level labels only than both sentence- and document-level labels, we compare the following two settings: (1) \Jointstruc, which uses additional manifestos with document-level supervision (\rile); and (2) \Jointsent, which uses manifestos with sentence-level supervision only. We vary the proportion of labeled documents at the sentence-level, from 10\% to 80\%, to study the effect under sparsely-labeled conditions. Note that 80\% is the maximum labeled training data under the cross-validation setting. In other cases, a subset (say 10\%) is randomly sampled for training. From \figref{JSoverJoint}, having more manifestos with document-level supervision demonstrates the advantage of semi-supervised learning, especially when the sentence-level supervision is sparse ($\le$ 40\%)--- \Jointstruc performs better than \Jointsent.

\subsection{Manifesto Position Re-ranking using PSL}
\label{sec:psl-results}

Finally, we present the results using PSL, which calibrates the overall manifesto position on the left--right spectrum, obtained using the joint-structured model (\Jointstruc). As we evaluate the effect of temporal dependency, we use  manifestos before 2008-09 for training (868 in total) and the later ones (until 2015, 136 in total) for testing. This test set covers one recent set of election manifestos for most countries, and two for the Netherlands, Spain and United Kingdom. To avoid variance in right-to-left ratio and the target variable (\logic{pos}, initialized using \Jointstruc) between the training and test sets, we build a stacked network \cite{fast2008stacked}, whereby we estimate values for the training set using cross-validation across the training partition, and estimate values for the test-set with a model trained over the entire training data. Note that we build the \Jointstruc model afresh using the chronologically split training set, and the parameters are tuned again using an 80-20 random split of the training set. For a consistent view of results for both the tasks (and stages), we provide micro-averaged results for sentence-classification with the competing approaches (from \tabref{sr}): \method{AE-NN} \cite{ALTW2017}, \method{Bi-LSTM(+up)}, and \Jointstruc. Results are presented in \tabref{sbt}, noting that the results for a given method will differ from earlier due to the different data split.


 \begin{table}
  \centering
  \begin{small}
  \begin{tabular}{ l c}
  \toprule
    Approach & F-measure\\
    \midrule
    \method{AE-NN}    &  0.31\\
    \method{Bi-LSTM(+up)}    &  0.36\\
    \Jointstruc &  \textbf{0.42}\\
    \bottomrule
  \end{tabular}
  \end{small}
  \caption{Micro-averaged F-measure for manifestos released after 2008-09. Best scores are given in bold.}
 \label{tab:sbt}
\end{table}

For the document-level regression task, we also evaluate other approaches based on manifesto similarity and automated scaling with sentence-level policy positions: 
\begin{itemize}
\item \textbf{Cross-lingual scaling (\method{CLS})}: A recent unsupervised approach for crosslingual political speech text scoring \cite{EACL}, based on TF-IDF weighed average word-embeddings to represent documents, and a graph constructed using pair-wise document similarity. Given two pivot texts (for left and right), label propagation approach is used to position other documents. 

\item \textbf{PCA}: Apply principal component analysis \cite{PCA} on the distribution of sentence-level policy positions (56 classes, without 000), and use the projection on its principal component to explain maximum variance in its sentence-level positions, as a latent manifesto-level position score.

\item \textbf{\Jointstruc}: We evaluate the scores obtained using  \Jointstruc, which we calibrate using PSL. 

\end{itemize}

We validate the calibrated position scores using both \rile and CHES\footnote{\url{https://www.chesdata.eu/}} scores. We use CHES 2010-14, and map the manifestos to the closest survey year (wrt its election date). CHES scores are used only for evaluation and not during training. We provide results in \tabref{psl} by augmenting features for the PSL model (\tabref{pslrule}) incrementally. We observed that the coalition-based feature, and polarity of sentences with its position information improves the overall ranking ($r$, $\rho$). Document similarity based relational feature provides only mild improvement (similarly to \newcite{Burford+:2015}), and temporal dependency provides further improvement against CHES. That is, combining content, network and temporal features provides the best results. 

 \begin{table}
  \centering
  \begin{small}
  \begin{tabular}{ lc c c@{\;} c c }
  \toprule
  &  \multicolumn{2}{c}{\rile} && \multicolumn{2}{c}{CHES} \\
    \cmidrule{2-3}
    \cmidrule{5-6}
  & $r$ & $\rho$ && $r$ & $\rho$\\
    \midrule
    \method{CLS}    &  0.11 & 0.10 && 0.09 & \pos0.07\\
    \method{PCA}    & 0.26 & 0.17 && 0.01  & $-$0.02 \\
    \Jointstruc  & 0.46 & 0.42&& 0.42 & \pos0.42\\
    \method{PSL$_\text{coal}$} & 0.51 & 0.45 && 0.49 & \pos0.45\\
    \method{PSL$_\text{coal + esim}$} & 0.52 & 0.47 && 0.50 & \pos0.46\\
    \method{PSL$_\text{coal + esim + ploc}$} & \textbf{0.54} & 0.56 && 0.53 & \pos0.56\\
    \method{PSL$_\text{coal + esim + ploc + temp}$} & \textbf{0.54} & \textbf{0.57} && \textbf{0.55} & \pos\textbf{0.61}\\
    \bottomrule
  \end{tabular}
  \end{small}
  \caption{Manifesto regression task using the two-stage approach. Best scores are given in bold.}
 \label{tab:psl}
\end{table}

\section{Conclusion and Future Work}
This work has been targeted at both fine- and coarse-grained manifesto text position analysis. We have proposed a two-stage approach, where in the first step we use a hierarchical multi-task deep model to handle the sentence- and document-level tasks together. We also utilize additional information on label structure, to augment an auxiliary structured loss. Since the first step places the manifesto on the left--right spectrum using text only, we leverage context information, such as coalition and temporal dependencies to calibrate the position further using PSL. We observed that: (a) a hierarchical bi-LSTM model performs best for the sentence-level classification task, offering a 10\% improvement over the state-of-art approach \cite{ALTW2017}; (b) modeling the document-level task jointly, and also augmenting the structured loss, gives the best performance for the document-level task and also helps the sentence-level task under sparse supervision scenarios; and (c) the inclusion of a calibration step with PSL provides significant gains in performance against both \rile and CHES, in the form of an increase from $\rho = 0.42$ to 0.61  wrt CHES survey scores. 

There are many possible extensions to this work, including: (a) learning multilingual word embeddings with domain information; and (b) modeling other policy related scores from text, such as ``support for EU integration''.

\section*{Acknowledgements}
We thank the anonymous reviewers for their insightful
comments and valuable suggestions. This
work was funded in part by the Australian Government
Research Training Program Scholarship, and
the Australian Research Council. 
\bibliography{naaclhlt2018}
\bibliographystyle{acl_natbib}
\end{document}